\documentclass{article}
\usepackage{spconf,amsmath,amssymb,graphicx}
\usepackage{booktabs,multirow}
\usepackage{enumitem}
\usepackage[table]{xcolor}
\definecolor{baselinebg}{RGB}{245,243,239}
\definecolor{pactbg}{RGB}{242,250,255}
\usepackage{microtype}
\usepackage{url}
\usepackage{needspace}
\setlist{nosep,leftmargin=14pt}
\newcommand{\ci}[1]{\hspace{0.25em}\textsubscript{\textcolor{black!55}{#1}}}
\newcommand{\taskdelta}[1]{\hspace{0.2em}\makebox[2.7em][l]{\textsubscript{\textcolor{black!55}{\ensuremath{#1}}}}}
\newcommand{\taskheading}[1]{\makebox[0pt][r]{\makebox[2.5em][c]{#1}}\taskdelta{}}
\newcommand{\deltacolhead}[1]{\makebox[0pt][r]{\makebox[2.5em][c]{$\Delta$}}\phantom{\ci{#1}}}

\title{EXACT: Executable Answer Contracts for Tool-Augmented \\ Medical Visual Reasoning}
\name{\fontsize{11}{13}\selectfont Zhenyu Yi$^{1}$, Zhenhao Li$^{1}$, Zelin Liu$^{1}$, Jiaxuan Zhao$^{1}$, Junqiao Wang$^{1}$, Juntao Jiang$^{2}$, Yusong Sun$^{1}$, Lichi Zhang$^{1}$\sthanks{Corresponding author.}}
\address{$^{1}$School of Biomedical Engineering, Shanghai Jiao Tong University\\
$^{2}$College of Control Science and Engineering, Zhejiang University}

\begin{document}
\maketitle

\begin{abstract}
Reliable medical image reasoning requires connecting anatomical observations and quantitative evidence to the conclusions they support. While multimodal large language models (MLLMs) can leverage specialized imaging tools to acquire such evidence, free-form tool use leaves the relationship between evidence and candidate answers implicit. We introduce \textbf{EXACT}, a training-free contract-execution harness that compiles candidate answers into a shared bounded program with explicit evidence requirements and option-specific predicates. These contracts guide targeted evidence acquisition, organize tool observations in a typed evidence ledger, and enable deterministic answer admission when one option is supported and all alternatives are refuted. On DeepTumorVQA, EXACT improves accuracy over direct inference by 3.76–8.16 percentage points across five general-purpose and medical backbones. It further achieves gains of 4.97 and 2.74 percentage points over same-tool ReAct on Qwen3.5-9B and Hulu-Med-7B, respectively. Fixed-evidence replay across three backbones independently demonstrates a 0.88–0.96-point improvement from deterministic admission. These results establish executable answer contracts as a practical mechanism for grounding CT visual question answering in verifiable evidence. Codes will be released upon publication.
\end{abstract}

\begin{keywords}
Multimodal Large Language Models, Evidence-grounded Reasoning, Computed Tomography
\end{keywords}

\suppressfloats[t]
\begin{figure}[t]
\centering
\includegraphics[width=\columnwidth]{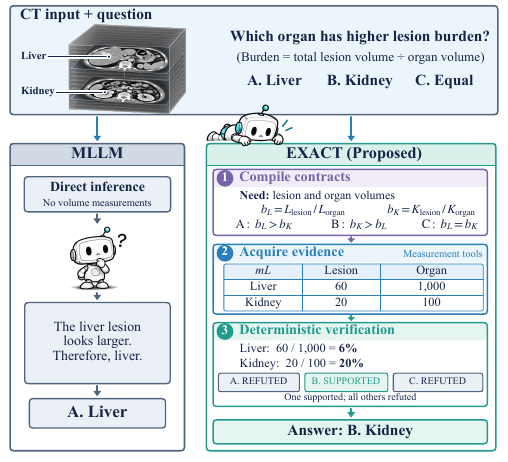}
\caption{Conceptual illustration of CT visual question answering (VQA), including the input, measurements, and responses. Evidence contracts guide targeted evidence acquisition and deterministic verification before answer admission.}
\label{fig:teaser}
\end{figure}

\section{Introduction}

Reliable medical image reasoning requires connecting anatomical observations and quantitative evidence to the conclusions they support. Multimodal large language models (MLLMs) have advanced medical image interpretation~\cite{xu2026lingshu,jiang2025hulu}, while recent volumetric approaches have improved cross-slice evidence aggregation, report-aligned reasoning, and efficient visual representations~\cite{yi2026brain,li2026mrl,yi2026carve}. For medical imaging visual question answering (VQA), a central challenge is to establish how acquired imaging evidence supports specific candidate answers, particularly when decisions depend on lesion measurements, tissue attenuation, or organ-level disease burden.

For example, a larger lesion does not necessarily imply a greater fraction of its organ is affected (Fig.~\ref{fig:teaser}). Comparing lesion burden requires measurements of both lesions and organs, bound to consistent anatomical targets and evaluated through explicit quantitative relationships. ReAct-based agents acquire evidence through reasoning-and-action loops~\cite{yao2022react}, while medical agents incorporate specialized imaging tools~\cite{li2024mmedagent,mao2026ct}. 3DMedAgent~\cite{wang20263dmedagent} aggregates tool outputs and slice-level evidence in structured memory for volumetric CT reasoning. Visual programming enables compositional perception and computation~\cite{gupta2023visual,suris2023vipergpt}, and structured perception cues expose information for downstream reasoning~\cite{janjua2026don}. Modular agents further support deterministic geometric reasoning over CT slices~\cite{abel2026modular}. Building on these advances, evidence-grounded CT VQA calls for an explicit connection between acquired facts, candidate-specific claims, and the conditions under which an answer can be accepted.

\begin{figure*}[!t]
\centering
\includegraphics[width=\textwidth]{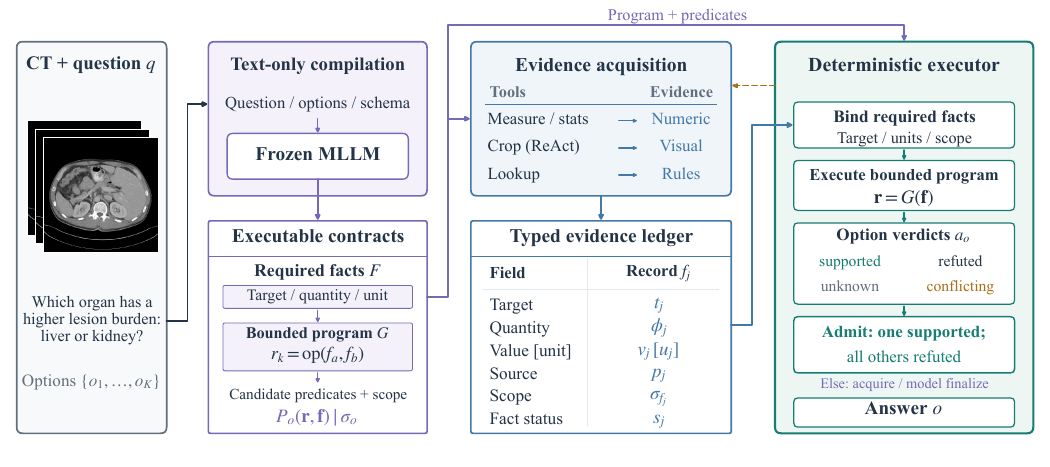}
\caption{Overview of EXACT. Text-only compilation produces a shared executable contract. Required facts guide acquisition; the executor binds typed evidence, executes a bounded program, and verifies candidate predicates before admission. Crops provide visual input to ReAct; dashed arrows denote budgeted reacquisition.}
\label{fig:method}
\end{figure*}

We introduce \textbf{EXACT}, a training-free contract-execution harness that makes candidate-level evidence requirements executable. EXACT compiles answer options into a shared bounded program with explicit evidence requirements, anatomical scope, and one verification predicate per candidate. These contracts guide targeted evidence acquisition and organize tool observations within a typed evidence ledger that preserves measurement units and provenance. A deterministic executor evaluates the resulting evidence against candidate predicates, admitting an answer when one option is supported and all alternatives are refuted. This design unifies evidence acquisition, quantitative computation, and answer verification within a contract-driven reasoning process.

We evaluate EXACT on DeepTumorVQA across five general-purpose and medical MLLMs. The complete harness improves accuracy over direct inference by 3.76--8.16 percentage points, with additional gains of 4.97 and 2.74 points over same-tool ReAct on Qwen3.5-9B and Hulu-Med-7B, respectively. Fixed-evidence replay across three backbones further demonstrates a 0.88--0.96-point improvement attributable to deterministic admission. Together, these results demonstrate the effectiveness of executable answer contracts for grounding CT VQA decisions in candidate-level verification.

\section{Method}
EXACT makes answer selection executable through a shared contract (Fig.~\ref{fig:method}). Required facts guide acquisition, a bounded program computes relevant quantities, and candidate predicates define acceptance conditions. A deterministic executor binds evidence and evaluates these conditions to admit an answer, request evidence, or invoke model finalization.

\begin{table*}[!t]
\centering
\caption{Accuracy (\%) on DeepTumorVQA v2 across five frozen backbones. Gray arrows: task-wise changes from Direct (pp). Overall is paired with its EXACT$-$Direct gain and gray paired 95\% CI; bold marks higher overall accuracy.}
\label{tab:main}
\smallskip
\small
\setlength{\tabcolsep}{5.7pt}
\begin{tabular}{>{\columncolor{white}}llrrrr>{\hspace{6pt}}rr}
\toprule
Backbone & Setting & \taskheading{Recog.} & \taskheading{Meas.} & \taskheading{VisR} & \taskheading{MedR} & \cellcolor{pactbg}\textbf{Overall} & \cellcolor{pactbg}\textbf{$\Delta$ vs Direct} \\
\cmidrule(lr){1-6}\cmidrule(lr){7-8}
\rowcolor{baselinebg}
\cellcolor{white} & Direct & 63.32\taskdelta{} & 26.30\taskdelta{} & 40.97\taskdelta{} & 50.13\taskdelta{} & 44.27 & -- \\
\rowcolor{pactbg}
\cellcolor{white}\multirow{-2}{*}{Qwen3.5-9B~\cite{qwen2026qwen35}} & EXACT & 58.48\taskdelta{\downarrow4.84} & 54.62\taskdelta{\uparrow28.32} & 42.65\taskdelta{\uparrow1.68} & 64.46\taskdelta{\uparrow14.33} & \textbf{52.43} & +8.16\ci{[6.80,9.57]} \\
\midrule
\rowcolor{baselinebg}
\cellcolor{white} & Direct & 54.37\taskdelta{} & 30.35\taskdelta{} & 33.54\taskdelta{} & 22.52\taskdelta{} & 35.01 & -- \\
\rowcolor{pactbg}
\cellcolor{white}\multirow{-2}{*}{Qwen3-VL-8B~\cite{bai2025qwen3}} & EXACT & 56.45\taskdelta{\uparrow2.08} & 50.67\taskdelta{\uparrow20.32} & 33.09\taskdelta{\downarrow0.45} & 23.88\taskdelta{\uparrow1.36} & \textbf{39.62} & +4.61\ci{[3.26,6.00]} \\
\midrule
\rowcolor{baselinebg}
\cellcolor{white} & Direct & 57.04\taskdelta{} & 26.89\taskdelta{} & 33.59\taskdelta{} & 25.88\taskdelta{} & 35.51 & -- \\
\rowcolor{pactbg}
\cellcolor{white}\multirow{-2}{*}{HealthGPT-Pro-8B~\cite{lin2025healthgpt}} & EXACT & 52.10\taskdelta{\downarrow4.94} & 51.95\taskdelta{\uparrow25.06} & 35.07\taskdelta{\uparrow1.48} & 30.50\taskdelta{\uparrow4.62} & \textbf{41.06} & +5.55\ci{[4.21,6.90]} \\
\midrule
\rowcolor{baselinebg}
\cellcolor{white} & Direct & 49.93\taskdelta{} & 29.66\taskdelta{} & 38.63\taskdelta{} & 24.46\taskdelta{} & 36.40 & -- \\
\rowcolor{pactbg}
\cellcolor{white}\multirow{-2}{*}{Lingshu-7B~\cite{xu2026lingshu}} & EXACT & 47.85\taskdelta{\downarrow2.08} & 52.40\taskdelta{\uparrow22.74} & 33.66\taskdelta{\downarrow4.97} & 32.81\taskdelta{\uparrow8.35} & \textbf{40.16} & +3.76\ci{[2.37,5.14]} \\
\midrule
\rowcolor{baselinebg}
\cellcolor{white} & Direct & 49.98\taskdelta{} & 29.41\taskdelta{} & 36.70\taskdelta{} & 32.18\taskdelta{} & 37.05 & -- \\
\rowcolor{pactbg}
\cellcolor{white}\multirow{-2}{*}{Hulu-Med-7B~\cite{jiang2025hulu}} & EXACT & 57.34\taskdelta{\uparrow7.36} & 39.05\taskdelta{\uparrow9.64} & 33.51\taskdelta{\downarrow3.19} & 44.83\taskdelta{\uparrow12.65} & \textbf{41.61} & +4.56\ci{[3.48,5.64]} \\
\bottomrule
\end{tabular}
\end{table*}

\subsection{Contract Compilation}
A frozen MLLM compiles the question, $K$ options, tool schemas, and available anatomical target names into a complete JSON contract in a unified schema. Compilation uses textual inputs without CT pixels or measurement values. The shared contract
\begin{equation}
C=\langle F,G,\{(\sigma_o,P_o)\}_{o=1}^{K}\rangle
\end{equation}
specifies required facts $F$, an ordered program $G$, and option predicates $P_o$ with claim scopes $\sigma_o$. Fact slots identify target, quantity, canonical unit, and scope. Program steps reference slots or preceding outputs, preserving dependencies between observations, derived quantities, and candidate claims.

Contracts contain up to four facts and eight steps drawn from arithmetic, ratios, percentages, comparisons, thresholds, intervals, numeric matching, and existence tests. A shared validator checks tool-expressible targets, operand references, option coverage, and the provenance of constants from the question, options, or retrieved rules before execution.

Numeric options are evaluated against the same acquired measurement. For measurement $v$ and candidate value $c$ in the same canonical unit, an approximate-match predicate requires $|v-c|\leq\max(0.02\max(|v|,|c|),0.05)$. This fixed tolerance accommodates numeric precision; overlapping matches remain inconclusive.

\subsection{Evidence Acquisition and Execution}
Contract requirements guide acquisition through organ crops, predicted measurements, segmentation statistics, and knowledge lookup~\cite{chen2026deeptumorvqa,wasserthal2023totalsegmentator}. Missing numeric facts map to measurement or segmentation requests, and retrievable rule gaps map to knowledge queries. Deduplicated requests let options reuse observations. Crops supply ReAct with visual input; numeric slots require quantitative tool outputs.

Each observation is represented in a typed evidence ledger as
\begin{equation}
f_j=\langle t,\phi,v,u,s,p,\sigma_f\rangle,
\end{equation}
with target $t$, quantity $\phi$, value $v$, unit $u$, status $s$ (available, unresolved, or visual), provenance $p$, and scope $\sigma_f$. Scope determines which claims a fact can support: a generic organ lesion count, for example, does not establish a specific pathology. Canonical units are assigned by requested quantity under the tools' unit conventions.

The executor binds required facts by target, quantity, and unit, checks their claim scope, and evaluates $\mathbf{r}=G(\mathbf{f})$ in dependency order. Derived quantities retain references to their source facts. Missing facts, incompatible units, invalid arithmetic, or failed scope checks yield \emph{unknown}; disagreeing duplicate values yield \emph{conflicting}. Unknown evidence alone cannot refute a candidate.

\subsection{Adjudication and Fallback}
Predicates assign each option a state $a_o$: \emph{supported}, \emph{refuted}, \emph{unknown}, or \emph{conflicting}. Admission requires one supported candidate and every alternative refuted:
\begin{equation}
a_{o^*}=\text{\emph{supported}}\ \wedge\ \forall o\ne o^*,\ a_o=\text{\emph{refuted}}.
\label{eq:admit}
\end{equation}
Thus, unknown or conflicting alternatives, multiple supported candidates, and all-refuted outcomes remain inconclusive. Actionable evidence gaps trigger budgeted acquisition and re-execution. When no further request is available, the model finalizer selects an answer from a text evidence digest without an injected verdict.

Unsupported contracts or compilation and validation failures invoke ordinary ReAct, where the MLLM can inspect crop images and integrate tool observations. Compilation, acquisition, and finalization share the episode budget. Qualitative visual claims and unresolved evidence use model-based selection.

\section{Experiments}
Across five frozen backbones, we evaluate EXACT's complete-system accuracy and interaction cost, then examine contract execution and deterministic admission through component ablations and fixed-evidence replay.

\subsection{Experimental Setup}
\noindent\textbf{Data and Backbones.} We use the DeepTumorVQA v2~\cite{chen2026deeptumorvqa} test set: 10,000 questions from 991 CT volumes spanning recognition, measurement, visual reasoning, and medical reasoning. Backbones include general-purpose Qwen3.5-9B~\cite{qwen2026qwen35} and Qwen3-VL-8B~\cite{bai2025qwen3}, and medical HealthGPT-Pro-8B (Qwen3-VL-based)~\cite{lin2025healthgpt}, Lingshu-7B~\cite{xu2026lingshu}, and Hulu-Med-7B~\cite{jiang2025hulu}. All weights remain frozen.

\noindent\textbf{Comparison Protocol.} Direct answers from a whole-volume montage without tools. Tool-only ReAct and EXACT share organ crops, knowledge lookup, and statistics and measurements from DeepTumorVQA's TotalSegmentator~\cite{wasserthal2023totalsegmentator} prediction cache. Both share decoding and context settings, with six model calls, eight tool calls, and four crops per episode. Inputs exclude gold masks, oracle measurements, answer labels, and subtype metadata.

\noindent\textbf{Metrics.} Overall accuracy pools all questions, counting invalid outputs as incorrect; Macro Avg. weights backbones equally. Paired 95\% CIs use 10,000 CT-volume cluster bootstrap resamples to preserve within-volume dependence.

\begin{table}[!t]
\centering
\caption{Ablations on Qwen3.5-9B~\cite{qwen2026qwen35}. Accuracy (\%); $\Delta$ versus Tool-only (pp). Compiled execution disables scope checks and reacquisition feedback.}
\label{tab:execution}
\smallskip
\small
\setlength{\tabcolsep}{11.8pt}
\begin{tabular}{lrr}
\toprule
Setting & Accuracy & \deltacolhead{[$-$0.07,1.47]} \\
\midrule
\rowcolor{baselinebg}
Tool-only ReAct & 47.46 & -- \\
Contract hint & 48.16 & +0.70\ci{[$-$0.07,1.47]} \\
Compiled execution & 52.19 & +4.73\ci{[3.70,5.78]} \\
Without scope checks & 52.26 & +4.80\ci{[3.81,5.82]} \\
Without feedback & 52.53 & +5.07\ci{[4.07,6.09]} \\
\rowcolor{pactbg}
EXACT (Ours) & 52.43 & +4.97\ci{[3.96,6.00]} \\
\bottomrule
\end{tabular}
\end{table}

\begin{table}[!t]
\centering
\caption{Fixed-evidence replay. On/Off: deterministic admission enabled/disabled. Accuracy (\%) and On$-$Off differences (pp) use the full test set; $n$: interventions.}
\label{tab:admission}
\smallskip
\small
\setlength{\tabcolsep}{5.9pt}
\begin{tabular}{lrrrr}
\toprule
Backbone & $n$ & On & Off & \deltacolhead{[0.66,1.26]} \\
\midrule
Qwen3.5-9B~\cite{qwen2026qwen35} & 815 & 52.43 & 51.55 & +0.88\ci{[0.67,1.10]} \\
Qwen3-VL-8B~\cite{bai2025qwen3} & 1,175 & 39.62 & 38.66 & +0.96\ci{[0.66,1.26]} \\
Hulu-Med-7B~\cite{jiang2025hulu} & 790 & 41.61 & 40.70 & +0.91\ci{[0.63,1.19]} \\
\bottomrule
\end{tabular}
\end{table}

\subsection{Main Results}

\begin{figure}[!t]
\centering
\includegraphics[width=\columnwidth]{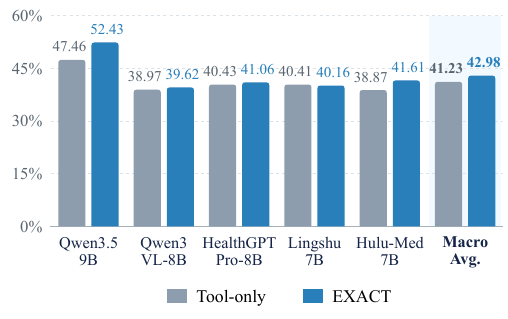}
\caption{Tool-only and EXACT accuracy (\%) with identical tools and budgets. Macro Avg. equally weights five backbones.}
\label{fig:tools}
\end{figure}

\noindent\textbf{Complete-System Performance.}
EXACT consistently outperforms Direct by 3.76--8.16 pp across all backbones, with paired CIs above zero (Table~\ref{tab:main}). Qwen3.5-9B achieves 52.43\% accuracy. Gains are particularly pronounced on measurement tasks, where accuracy rises from 26.30\% to 54.62\% on Qwen3.5-9B and from 26.89\% to 51.95\% on HealthGPT-Pro. This pattern is consistent with EXACT's quantitative design: shared measurements feed explicit computations and candidate predicates. Recognition also improves on Qwen3-VL-8B and Hulu-Med-7B. Table~\ref{tab:main} evaluates the complete harness; Fig.~\ref{fig:tools} holds tools fixed to assess contract-driven reasoning.

\noindent\textbf{Matched-Tool Comparison.}
Under identical tools and budgets (Fig.~\ref{fig:tools}), EXACT gains 4.97 pp over Tool-only on Qwen3.5-9B and 2.74 pp on Hulu-Med-7B, raising five-backbone mean accuracy from 41.23\% to 42.98\%.

\noindent\textbf{Interaction Cost.}
EXACT reduces both model and tool calls across all five backbones. On Qwen3.5-9B, it averages 3.70 model and 2.88 tool calls per question, compared with 4.89 and 4.43 for Tool-only, respectively. Counts include compilation and finalization.

\noindent\textbf{Case-Disjoint Evaluation.}
A 400-question subset from 39 test volumes was used for development. Excluding those volumes leaves 9,600 questions from 952 CT volumes, on which EXACT maintains gains of 3.59--8.18 pp over Direct across five backbones. Qwen3.5-9B retains a matched-tool gain of 4.82 pp (95\% CI: [3.78, 5.88]). These results score frozen predictions on cases excluded from development.

\subsection{Contract Execution and Answer Admission}

\noindent\textbf{Contract Execution.}
On Qwen3.5-9B, compiled contracts gain 4.73 pp over Tool-only, versus 0.70 pp when used as ReAct hints (Table~\ref{tab:execution}). Contract hints supply plans without predicate execution. EXACT achieves 52.43\%, versus 52.26\% without scope checks and 52.53\% without feedback. Scope checks enforce claim eligibility, while feedback enables budgeted reacquisition. Across five backbones, 51.4--65.7\% of questions reach valid-contract execution, and 7.9--13.9\% satisfy the unique-answer admission criterion (Eq.~\ref{eq:admit}). Execution guides acquisition even when the evidence does not uniquely separate all candidates.

\noindent\textbf{Fixed-Evidence Admission.}
Fixed-evidence replay replaces admission with a verdict-blind backbone finalizer, preserving contracts and observations (Table~\ref{tab:admission}). Accuracy is scored over all 10,000 questions, with predictions outside the intervention cases held fixed. Across three backbones, admission improves full-test accuracy by 0.88--0.96 pp, with paired CIs above zero. On Qwen3.5-9B, it corrects 104 answers while introducing 16 errors across 815 interventions. Online Qwen3.5 traces also show gains on inconclusive episodes, with acquisition allowed to differ between runs. Replay isolates answer selection under fixed evidence.

\section{Conclusion}
EXACT unifies targeted evidence acquisition, deterministic verification, and answer admission through executable contracts. Cross-backbone gains, matched-tool comparisons, and fixed-evidence replay establish contract execution and deterministic admission as effective mechanisms for evidence-grounded medical imaging VQA.
\section{Compliance with ethical standards}
No additional ethical approval was required for this study.

\section{Acknowledgments}
This research is supported by the National Natural Science Foundation of China (Grant No. 62471288). The authors declare no conflicts of interest. The code will be made publicly available upon acceptance. For AI usage, we use ChatGPT to polish the manuscript and check for grammatical errors.

\bibliographystyle{IEEEbib}
\bibliography{refs}
\end{document}